\documentclass[twoside,leqno,twocolumn]{article}

\usepackage[letterpaper]{geometry}

\usepackage{ltexpprt}
\usepackage{hyperref}
\usepackage{float}

\usepackage{url}            
\usepackage{booktabs}       
\usepackage{amsfonts}       
\usepackage{nicefrac}       
\usepackage{microtype}      
\usepackage{xcolor}         
\usepackage{multirow}
\usepackage{makecell}
\usepackage[english]{babel}
\usepackage{amsthm}
\usepackage{weiwAlgorithm}
\usepackage{amsmath}
\usepackage{subfig}
\usepackage{graphicx} 
\usepackage{setspace}

\DeclareMathOperator*{\argmax}{arg\,max} 
\begin{document}

\newtheorem{definition}{Definition}[]
\newtheorem{assumption}{Assumption}[]
\newtheorem{corollary}{Corollary}[]

\newcommand{\LESA}{LEIP\xspace}

\newcommand\relatedversion{}
\renewcommand\relatedversion{\thanks{The full version of the paper can be accessed at \protect\url{https://arxiv.org/abs/1902.09310}}} 

\title{Label Shift Estimation With Incremental Prior Update}

\author{Yunrui Zhang\thanks{University of New South Wales, School of Computer Science and Engineering, Australia Email: \{yunrui.zhang, g.batista, salil.kanhere\} @unsw.edu.au \\ Corresponding author: Yunrui Zhang}
\and Gustavo Batista \footnotemark[1]
\and Salil S. Kanhere \footnotemark[1]}

\date{}

\maketitle






\begin{abstract}
   An assumption often made in supervised learning is that the training and testing sets have the same label distribution. However, in real-life scenarios, this assumption rarely holds. For example, medical diagnosis result distributions change over time and across locations; fraud detection models must adapt as patterns of fraudulent activity shift; the category distribution of social media posts changes based on trending topics and user demographics. In the task of label shift estimation, the goal is to estimate the changing label distribution $p_t(y)$ in the testing set, assuming the likelihood $p(x|y)$ does not change, implying no concept drift. In this paper, we propose a new approach for post-hoc label shift estimation, unlike previous methods that perform moment matching with confusion matrix estimated from a validation set or maximize the likelihood of the new data with an expectation-maximization algorithm. We aim to incrementally update the prior on each sample, adjusting each posterior for more accurate label shift estimation. The proposed method is based on intuitive assumptions on classifiers that are generally true for modern probabilistic classifiers. The proposed method relies on a weaker notion of calibration compared to other methods. As a post-hoc approach for label shift estimation, the proposed method is versatile and can be applied to any black-box probabilistic classifier. Experiments on CIFAR-10 and MNIST show that the proposed method consistently outperforms the current state-of-the-art maximum likelihood-based methods under different calibrations and varying intensities of label shift.
\end{abstract}

\section{Introduction}
A ubiquitous assumption in supervised learning is that the training and testing sets have the same class distribution. However, label shifts are common in real-life scenarios, meaning the class prevalence differs between training and testing time. For example, a machine learning model trained to predict a disease based on patient data from a specific hospital may encounter the same diseases but with different prevalences when deployed in a different hospital or region with varying seasons and demographics.

We assume the existence of two probability distributions. Data from the \textit{source} distribution $p_s$ is available during training time, while the classifier is deployed in a \textit{target} distribution $p_t$ that could be different to the source distribution $p_s$. For a feature set $\mathbf{x}$ and a label $y$, a classifier under Bayes' theorem computes a posterior label probability through $p_t(y|\mathbf{x}) = \frac{p_t(\mathbf{x}|y)p_t(y)}{p_t(\mathbf{x})}$. In the context of label shift, we assume that the likelihood $p_t(x|y)$ does not change from the source distribution, implying the characteristics of the features remain the same for each class label. In contrast, the prior $p_t(y)$ and the overall feature distribution $p_t(x)$ may change concerning the source distribution.  Label shift estimation involves estimating the class prevalence in the target distribution, $p_t(y)$. The estimated $p_t(y)$ can be used in tasks such as quantification~\cite{gonzalez_review_2017}, positive and unlabelled learning (PU-learning)~\cite{elkan2008learning} and imbalanced semi-supervised learning \cite{kim2020distribution, wei2021crest, zhu2023generalized}. Alternatively, we can adapt or retrain the classifier with importance weighting \cite{fang2020rethinking} to improve its performance in the target distribution~\cite{saerens_adjusting_2002}. 


 
Two post-hoc approaches dominate label shift estimation: the confusion matrix approach~\cite{azizzadenesheli_regularized_2019, lipton_detecting_2018, mclachlan2005discriminant} and the expectation-maximization (EM) approach~\cite{saerens_adjusting_2002}. In previous work~\cite{alexandari2020maximum, garg_unified_2020}, the EM approach~\cite{saerens_adjusting_2002} with bias-corrected calibration has shown to be the best-performing approach for label shift estimation, significantly outperforming confusion matrix approaches~\cite{azizzadenesheli_regularized_2019, lipton_detecting_2018}. Both methods can be generalized as distribution matching in latent space. At the same time, the EM algorithm would require the latent space generated by a multi-class complete calibrated classifier~\cite{garg_unified_2020}.

In this chapter we proposes \LESA, a novel approach for \textbf{L}abel shift \textbf{E}stimation with \textbf{I}ncremental \textbf{P}rior update. Unlike the existing methods for post-hoc labels shift estimation that performs moment matching with confusion matrix estimated from the validation set or maximizes the likelihood of the new data with expectation-maximization algorithm, \LESA takes the alternative objective to adjust the testing set prediction by performing incremental prior update on each sample to adjust the posterior which would lead to a more accurate label shift estimation. \LESA assumes the underlying probabilistic classifier is relatively accurate, leading to a long-tailed distribution of the top one probabilistic output in which the top-ranked instances are correctly classified. These assumptions are either intuitive for any modern probabilistic classifiers or prerequisites for the probabilistic classifiers for the task of label shift estimation to be meaningful. We show that the proposed algorithm relies on a weaker notion of calibration than the complete calibrated classifier required by the EM approach. Due to the non-iterative approach and by operating solely on the probabilistic output, \LESA has a linear time complexity and is highly scalable. Like previous post-hoc label shift estimation methods, \LESA is compatible with any black-box classifier that produces probabilistic outputs. We evaluate \LESA on MNIST and CIFAR-10, and the empirical results show \LESA consistently outperforms the current label shift estimation state-of-the-art method EM under different calibrations and different levels of label shift.

\section{Problem Setup}

In supervised tasks, a dataset has a feature set $\mathbf{x}\in\mathbb{R}^{d}$ and labels $y\in\mathcal{Y}$, being $Y$ discrete set with $m$ possible values, \textit{i.e.}, $Y=\{y_1,\ldots,y_m\}$. A probabilistic classifier $\mathcal{F}$ provides a score to a sample $\mathbf{x}$ from a source distribution $s$ according to Bayes' theorem:

\[\hat{p}_s(y|\mathbf{x}) = \frac{p_s(\mathbf{x}|y)p_s(y)}{p_s(\mathbf{x})}\]

After the classifier deployment, suppose we encounter a target distribution $t$ where $p_t(y) \neq p_s(y)$, leading to a change in the overall feature distribution, $p_t(\mathbf{x}) \neq p_s(\mathbf{x})$. The classifier $\mathcal{F}$ gives the new posterior:

\[\hat{p}_t(y|\mathbf{x}) = \frac{p_t(\mathbf{x}|y)p_t(y)}{p_t(\mathbf{x})}\]

We assume no concept drift in the target distribution, implying that $p_t(\mathbf{x}|y) = p_s(\mathbf{x}|y)$. Thus, we refer to both as $p(\mathbf{x}|y)$ in this paper. For the task of label shift estimation, the target is to estimate the $p_t(y)$, the target label distribution. Alternatively, we can also report $w = \frac{p_t(y)}{p_s(y)}$, \textit{i.e.} the ratio between the target and the source class prevalences. Several articles report and compare the ``weight'' $w$ as it can be directly used to update the class posteriors from source to target distribution or as importance weighting to retrain the classifier.

Most existing classifiers, including deep neural networks, are uncalibrated. Therefore, the estimated posterior class probabilities $\hat{p}_s(y|\mathbf{x})$ and $\hat{p}_t(y|\mathbf{x})$ are not an accurate estimation of the true posteriors $p_s(y|\mathbf{x})$ and $p_t(y|\mathbf{x})$. These classifiers often provide over-confident posterior estimates~\cite{guo_calibration_2017, platt1999probabilistic, silva_filho_classifier_2023} and need to be calibrated to bring them closer to the real posteriors. Three main notions exist for multiclass calibration corresponding to different calibration levels \cite{kull_beyond_2019}. The strongest is the complete calibration. The other weaker calibration forms, confidence and classwise, are also commonly used. They are defined as:

\begin{definition} \textbf{Complete Calibration}. A classifier $\mathcal{F}$ that outputs $\{\hat{p}(y_i|\mathbf{x})\}_{i=1}^m$ is complete calibrated if  $\hat{p}(y_i|\mathbf{x}) = p(y_{i}|\mathbf{x})$ for all classes $y_i$.
\label{CAL:1}

\end{definition}
\begin{definition} \textbf{Confidence Calibration}.  A classifier $\mathcal{F}$ is confidence calibrated if $\argmax_y \hat{p}(y|\mathbf{x}) = \argmax_y p(y|\mathbf{x})$ and $\max \hat{p}(y|\mathbf{x}) = \max p(y|\mathbf{x})$. 
\label{CAL:2}

\end{definition}
\begin{definition} \textbf{Classwise Calibration}. A classifier $\mathcal{F}$ is classwise calibrated for class $y_i$ if $\hat{p}(y_i|\mathbf{x}) = p(y_{i}|\mathbf{x})$.
\label{CAL:3}
\end{definition}

\section{Related Work}

The maximum likelihood and confusion matrix approaches are the two post-hoc label shift estimation methods considered state-of-the-art. Both of the approaches operate on the posterior class probabilities generated by any black-box classifier that provides a probabilistic output instead of performing feature space distribution matching \cite{6279353, zhang2013domain, guo2020ltf}, making them scalable and versatile.

Saerens et al.~\cite{saerens_adjusting_2002} proposed the maximum likelihood method for label shift estimation. The method is based on the renowned expectation maximization algorithm. The following equation computes the posteriors in the expectation (E) step in the iteration $l$:

\begin{equation}
\hat{p}^{(l)}(y_i|\mathbf{x}_k)=\frac{\frac{\hat{p}^{(l)}(y_i)}{p_s(y_i)}\hat{p}_s(y_i|\mathbf{x}_k)}{\sum\limits_{j=1}^{m}\frac{\hat{p}^{(l)}(y_j)}{p_s(y_j)}\hat{p}_s(y_j|\mathbf{x}_k)}
\label{EM:1}
\end{equation}

Equation~\ref{EM:1} is computed for each test sample $\mathbf{x}_k$ and class $y_i$. In addition, the class prevalences $\hat{p}^{(l+1)}(y)$ are updated in the maximization (M) step by:

\begin{equation}
\hat{p}^{(l+1)}(y_{i}) = \frac{1}{N}\sum\limits_{k=1}^{N}\hat{p}^{(l)}(y_i|\mathbf{x}_k)
\label{EM:2}
\end{equation}

Where $N$ is the number of test set instances. The E and M steps continue until convergence when the last class prevalence is returned as estimated label distribution. The original paper \cite{saerens_adjusting_2002, latinne2001adjusting} initializes $\hat{p}^{(0)}(y_i)$ and $p(y_i)$ with the source label distribution. However, more recent work suggests that using the average probabilistic output of the validation set $\mathbb{E}_{i}[p_s(y_i|\mathbf{x})]$ yields better results~\cite{alexandari2020maximum}. For the rest of this paper, we refer to the expectation-maximization algorithm simply as EM. One underlying assumption of EM for its convergence to the true $p_t(y)$ is the classifier needs to be completely calibrated, and a lesser form of calibration is insufficient~\cite{garg_unified_2020}. 

The confusion matrix-based approach has been one of the oldest approaches for label shift estimation~\cite{mclachlan2005discriminant}. It is also used by Saerens et al.~\cite{saerens_adjusting_2002} as a baseline method. More recently, it has been popularized by Lipton et al.~\cite{lipton_detecting_2018} and Azizzadenesheli et al.~\cite{azizzadenesheli_regularized_2019} that propose BBSL and RLLS respectively, with RLLS being an improved version of BBSL with regularization. The confusion matrix-based method starts by estimating the confusion matrix $C_{\hat{y},y}$ on the validation set. The estimated weight between the new label distribution and the source is calculated by $\hat{w}=C^{-1}_{\hat{y},y}\hat{u}_{\hat{y}}$, where $\hat{u}_{\hat{y}}$ is the mean of the test set probabilistic output for all classes.

Empirically, EM with bias-corrected calibration methods is the best-performing method~\cite{alexandari2020maximum}, consistently outperforming both BBSL and RLLS. Garg et al.~\cite{garg_unified_2020} generalize the EM and confusion matrix-based methods under the same distribution matching framework. The difference is that the confusion matrix approach obtains the calibrated predictor via the confusion matrix, while the EM relies on a separate calibrator. 

One key assumption for post-hoc label shift estimation that is introduced in the original EM paper but rarely mentioned in the later work is that the posterior probabilities calculated from the test set by the classifier are reasonably well approximated, implying the classifier need to be reasonably accurate.~\cite{saerens_adjusting_2002}.

\section{\LESA}

This paper proposes \LESA, a novel approach for label shift estimation with incremental prior updates. \LESA takes an alternative per-sample prior update approach that adjusts the prediction of each sample, which would lead to a more accurate label shift estimation. This approach differs from the EM method, which maximizes the likelihood of observing the test set, or the confusion matrix-based approach, which predicts the prevalence according to the estimated confusion matrix on the validation set.

\LESA is motivated by obtaining a more accurate label shift estimation by adjusting and correcting samples. It is also based on the observation that highly confident samples are unlikely to be incorrect for a calibrated probabilistic classifier. In addition to the unchanging likelihood between the source and target distributions, which the label shift estimation task relies on, \LESA makes two assumptions about the probabilistic classifier.

\begin{assumption}
\textbf{Accurate Classifier} The classifier $\mathcal{F}$ provides predicted probabilities that are sufficiently close to the observed probabilities, indicating accurate probabilistic estimates.
\label{Assumption3}
\end{assumption}

\begin{assumption}
\textbf{Long-tailed Probabilities Output} For a probabilistic classifier $\mathcal{F}$, the distribution of the top-one predicted probabilities, $max_y(\hat{p}_s(y|\mathbf{x}))$, exhibits a long-tailed behavior on the test set. This implies that $\mathcal{F}$ assigns high confidence to a majority of samples, with fewer samples receiving significantly lower confidence scores.
\label{Assumption1}
\end{assumption}

Assumption \ref{Assumption3} assumes that the posterior provided by the classifier is well approximated, implying the classifier is relatively accurate and calibrated. This assumption is an assumption that EM also explicitly relies on~\cite{saerens_adjusting_2002} and arguably a fundamental assumption underlying all post-hoc label shift estimation methods that operate on the posterior generated by probabilistic classifiers. For a classifier that cannot provide a relatively well-approximated posterior, any adjustment on the posterior or estimation based on the posterior would be unjustified and not robust. 

As for Assumptions~\ref{Assumption1}, it could be seen as a derivation of Assumption~\ref{Assumption3}. For a relatively accurate and well-calibrated probabilistic classifier, as measured by Expected Calibration Error (ECE) \cite{guo_calibration_2017}, the top-1 prediction probabilities exhibit a long-tailed distribution due to the characteristics of ECE. This assumption is intuitive based on the definition of ECE and has been empirically observed across a wide range of probabilistic models that are relatively accurate and well-calibrated.

By combining Assumption~\ref{Assumption3} and Assumption~\ref{Assumption1}, we can reach the observation that a classifier satisfying Assumptions~\ref{Assumption3} and \ref{Assumption1} being accurate and well-calibrated, while also exhibiting a long-tailed distribution in its top-1 probabilistic outputs. When tested on any batch of test samples, the subset with high top-1 probabilities should be highly accurate, with few to no misclassifications, and this subset represents a significant portion of the training set. \LESA builds on this observation by using the class distribution of this accurate subset as the starting point for label shift estimation in the algorithm.

The overall procedure of \LESA is as follows. For each of the top label probabilistic outputs $\hat{q}_{k} = \max(\hat{p}_t(y|\mathbf{x}_{k}))$ of classifier $\mathcal{F}$ on the testing set $\mathcal{X}_t$ with unknown label distribution $p_t(y)$. \LESA start by using a threshold $\tau$ that is to select a subset of samples $\mathcal{X}_{A} \in \mathcal{X}_t$ such that $\hat{q}_{k} \ge \tau$. And a corresponding set of top label predictions $L_{A}$ that is $argmax(\hat{p}_t^{a}(y|\mathbf{x})) \in L_{A}$.

We take $\mathcal{X}_{A}$ as the set of samples that the prediction is confident. Empirically, with appropriate $\tau$ and a classifier that meets the Assumption \ref{Assumption3} and Assumption \ref{Assumption1}, the $\mathcal{X}_{A}$ set would have much higher accuracy than the overall accuracy, and empirically this set of samples has the accuracy that is close to one. Also as the $\hat{q}_{s}$ has a long tail distribution according to Assumption \ref{Assumption1}, $\mathcal{X}_{A}$ should comprise the majority within $\mathcal{X}$. The class distribution $\hat{p}^{(a)}(y)$ of $L_{A}$ can be seen as an approximation of  $\hat{p}_{\tilde{t}}(y)$ and will be used as the starting point of updating the priors in the $\mathcal{X}_{B}$ that is $\mathcal{X} \setminus \mathcal{X}_{A}$ the set of the samples that each sample $\hat{q}_{k}^{(b)} < \tau$. 

The set $\mathcal{X}_{B}$ is sorted into a descending ordered list $\mathcal{B}$ that $\mathcal{B} = \{\hat{q}_{1}^{(b)} \geq \hat{q}_{2}^{(b)} \geq \ldots \geq \hat{q}_{n}^{(b)}\}$ so that the incremental prior update can be performed in an ascending confidence manner. By sorting the $\mathcal{X}_{B}$ and performing the incremental prior update in the ascending order of confidence \LESA would incorporate the samples that we have higher confidence first and avoid the accumulation of error.

Then for each sample $\mathbf{x}_{k}$ in  $\mathcal{B}$ and the corresponding probabilistic output $\hat{p}^{(b)}(y|\mathbf{x}_{k})$ in the ordered list $\mathcal{B}$. An updated posterior $\hat{p}^{(c)}(y|\mathbf{x}_{k})$ is obtained by Equation \ref{LESA:1}, a derivation of the Bayes' theorem with the assumption that the likelihood stays the same in the target and source distributions. This equation is similar to the E step of the EM algorithm \cite{saerens_adjusting_2002}, but in a sample-wise manner without aggregation. Where the ${p_s(y)}$ is the source label distribution obtained from the training or validation set, and the $\hat{p}^{(a)}(y)$ is the label distribution of $L_{A}$.

\begin{equation}
\label{LESA:1}
\hat{p}^{(c)}(y_{i}|\mathbf{x}_{k})=\frac{\frac{\hat{p}^{(a)}(y_{i})}{p_s(y_{i})}\hat{p}^{(b)}(y_{i}|\mathbf{x}_{k})}{\sum\limits_{j=1}^{m}\frac{\hat{p}^{(a)}(y_{j})}{p_s(y_{j})}\hat{p}^{(b)}(y_{j}|\mathbf{x}_{k})}
\end{equation}

\begin{equation}
\label{LESA:2}
L_{A} = L_{A} \cup \{argmax(\hat{p}^{(c)}(y|\mathbf{x}_{k}))\}
\end{equation}

For each of the obtained updated posterior $\hat{p}^{(c)}(y|\mathbf{x}_{k})$ the top label prediction $argmax(\hat{p}^{(c)}(y|\mathbf{x}_{k}))$ is taken and combined with $L_{A}$ as shown in Equation~\ref{LESA:2}. When updating the prior for each sample in $\mathcal{B}$, the $\hat{p}^{(a)}(y)$ in Equation \ref{LESA:1} is the class distribution of the updated $L_{A}$ from the prior step. 

After the prior update of every sample in $\mathcal{B}$ with Equation \ref{LESA:1}, the class distribution estimation $\hat{p}_{e}(y)$ is obtained by taking the label distribution of the final $L_{A}$. The $\hat{p}_{e}(y)$ is then used to perform prior update on all the $\hat{p}_t(y|\mathbf{x})$ for every sample in $\mathcal{X}_t$ by Equation \ref{LESA:1} to obtain the $\hat{p}_{\tilde{t}}(y|\mathbf{x})$ and the corresponding $\hat{q}_{\tilde{t}}$ from which we can obtain the $\hat{p}_{\tilde{t}}(y)$. The estimation for $p_{t}(y)$. The one last complete pass on the $\mathcal{X}_t$ is to perform the prior update on the samples that are filtered out by $\tau$ but $argmax(\hat{p}_{t}(y|\mathbf{x})) \neq argmax(p_{t}(y|\mathbf{x}))$, the samples that are incorrectly filtered out by $\tau$. Algorithm \ref{alg:fea_sel} details \LESA.

\begin{algorithm}[h!]
 \setstretch{1.25}
\KwIn{Classifier probabilistic output $\hat{p}_{t}(y|\mathbf{x})$,}
\hspace{31pt}Source distribution $p_{s}(y)$, \\
\hspace{31pt}Threshold $\tau$ \\
\KwOut{Estimated $\hat{p}_{\tilde{t}}(y)$} 
$ A \leftarrow \hat{p}_{t}(y|\mathbf{x})$ that $\max(\hat{p}_{t}(y_{i}|\mathbf{x})) \ge \tau$\\
$ L_{A} \leftarrow argmax(A)$ \\
$ \hat{p}^{(a)}(y) \leftarrow \hat{p}(y)$ of $L_{A}$ \\
$ B \leftarrow \hat{p}_{t}(y|\mathbf{x})$ that $\max(\hat{p}_{t}(y_{i}|\mathbf{x})) < \tau$\\
$ B$ = ReverseSort($B$)\\
\ForEach{$\hat{p}_{t}(y|\mathbf{x}_{k}) \in B$}{  
    Update $\hat{p}_{t}(y|\mathbf{x}_{k})$ according to Equation \ref{LESA:1} and obtain $\hat{p}_{c}(y|\mathbf{x}_{k})$\\
    Update $L_{A}$ and $\hat{p}^{(a)}(y)$ according to Equation \ref{LESA:2}
}
$\hat{p}_{e}(y) \leftarrow \hat{p(y)}$ of $L_{A}$

$ L_{t} \leftarrow \emptyset$\;
\ForEach{$\hat{p}_{t}(y|\mathbf{x}_{k}) \in \hat{p}_{t}(y|\mathbf{x}) $}{  
    Update $\hat{p}_{t}(y|\mathbf{x}_{k})$ according to Equation \ref{LESA:1} with $\hat{p}_{e}(y)$ as $\hat{p}^{(a)}(y)$ and obtain $\hat{p}_{c}(y|\mathbf{x}_{k})$\\
    $L_{t} \leftarrow L_{t} \cup \{argmax(\hat{p}_{c}(y|\mathbf{x}_{k}))\}$
}
$\hat{p}_{\tilde{t}}(y) \leftarrow$ $\hat{p(y)}$ of $L_{t}$\\
return $\hat{p}_{\tilde{t}}(y)$
\caption{\LESA}
\label{alg:fea_sel}
\end{algorithm}

\subsection{Selecting $\tau$}
For the threshold $\tau$, it needs to provide a reasonably large number of samples for us to use in estimating the starting point of \LESA while also leaving as many samples as possible that need to be corrected. We propose using the minimum recall among all classes calculated based on the validation set confusion matrix to achieve this.

\begin{corollary} For a classifier $\mathcal{F}$ with a probability confusion matrix $C_{\hat{y}, y}$ estimated from the validation set, where each column $C_{\cdot, y}$ sums to one. The minimum accuracy when predicting on a set of samples with an unknown label distribution is given by the minimum recall that is $ \min(\text{diag}(C_{\hat{y}, y})) $, assuming $C_{\hat{y}, y} = \hat{C}_{\hat{y}, y}$.
\label{corollary1}
\end{corollary}

Corollary \ref{corollary1} demonstrates that the minimum accuracy for a classifier predicting under label shift is the minimum recall across all classes, as estimated from the validation set. This holds under the assumption that the probability confusion matrix remains consistent between the validation and testing sets, a consequence of the no covariate shift assumption. In the context of label shift estimation, we assume that the likelihood $p(\mathbf{x}|y)$ remains unchanged during testing, implying no concept drift. Consequently, the recall for each class does not change with variations in the class distribution. Therefore, the minimum possible accuracy occurs when all testing set samples belong to the class with the lowest recall, making the accuracy equal to the minimum recall. Base on Corollary \ref{corollary1}, to select $\tau$ for \LESA we take classifier's top probabilistic output $\hat{q}_t = max(\hat{p}_t(y|\mathbf{x}))$ for $\mathcal{X}_t$. The $\tau$ is taken set by the sample value at the top n percentile of $\hat{q}_t$ in which $n$ being the $min(diag(C_{\hat{y}, y}))\times100$. 

Using the cutoff at minimum recall for $\tau$ ensures that the set $\mathcal{X}_{A}$ is close to 100\% accurate, as shown in Corollary \ref{corollary1}. Also, assuming a relatively accurate classifier, the recall shouldn't be excessively low, resulting in $X_{A}$ being reasonably large. When the classifier is significantly inaccurate for certain classes, alternative values for $\tau$ should be used to ensure a reasonably large $\mathcal{X}_{A}$, such as using the average recall of the classifier on the validation set. By setting $\tau$ in accordance with the validation set's confusion matrix, we reduce the need for $\tau$ as a hyperparameter and provide robust justification for its selection.

\subsection{The Effect of Calibration}
One main difference between \LESA to the EM and the confusion matrix-based methods is that the \LESA takes a per instance update measure and does not perform any aggregation on the posterior. Moreover, in the \LESA procedure for the updated posteriors $\hat{p}^{(c)}(y|\mathbf{x}_{k})$ as well as the posteriors for $\mathcal{X}_{A}$ only the top value label is taken for prior update and label distribution estimation. While for EM, the values of posteriors are taken by the mean in each iteration and need to be thoroughly calibrated. 

As \LESA does not aggregate the posterior classwise and only selects the top label after the prior update, it necessitates a weaker calibration notion than EM. This variant of confidence calibration only considers the label with the highest confidence rather than the confidence values themselves. Additionally, it requires the top label after the prior update, $argmax(\hat{p}_{(s)}(\mathbf{x}|y))$ to match the true top label after the prior update by Equation \ref{LESA:1}. This new notion of calibration is defined as follows.

\begin{definition} \textbf{Top label Calibration in Changing Prior} The classifier $\mathcal{F}$ is top label calibrated in changing prior if for any probabilistic output $\hat{p}(\mathbf{x}|y)$, and the prior updated posterior $\hat{p}_{t}(\mathbf{x}|y)$ that is prior updated with $p_{t}(y)$ with Equation \ref{LESA:1}, the $argmax(\hat{p}_{t}(\mathbf{x}|y)) = argmax(p_{t}(\mathbf{x}|y))$
\label{def:LESA_CAL}
\end{definition}

For Definition \ref{def:LESA_CAL}, it is clear to see that for $p_t(y) = \mathbb{R}^{m}$ and $\sum_{i=1}^{m} p_t(y_{i}) = 1$. The top label Calibration would require complete calibration to ensure the correctness of the top label in all possible new priors $p_t(y)$. However, for most $p_t(y)$ in the context of label shift, the Top label Calibration in Changing Prior holds without the posterior to be completely calibrated and thus is a weaker notion of calibration in comparison to complete calibration in Definition \ref{CAL:1} that EM is relied on.


\section{Experiments}
\begin{table*}[htbp]
\centering
\small
\setlength{\tabcolsep}{6.5pt}
\begin{tabular}{|cc|ccc|ccc|ccc|}
\hline
\makecell{Shift \\ Estimator}& \makecell{Calibration \\ Method} & \multicolumn{3}{c|}{$\alpha=0.1$} &\multicolumn{3}{|c|}{$\alpha=1$}&\multicolumn{3}{|c|}{$\alpha=10$}\\
\hline
\multicolumn{2}{|c}{Validation Size} & 1000 & 2500 & 4000& 1000 & 2500 & 4000 & 1000 & 2500 & 4000 \\
\hline
CC        & None & 53.2 & 50.1 & 49.6 & 9.78 & 9.23 & 9.09 & \textbf{1.41} & 1.30 & 1.29 \\
RLLS-Hard & None & 6.68 & \textbf{3.85} & \textbf{2.61} & 5.02 & \textbf{2.03} & 1.58 & 2.66 & 1.15 & 1.10 \\
RLLS      & None & \textbf{6.07}  & 5.27  & 2.60  & 5.42 & 2.55 & \textbf{1.22} & 3.38 & 1.62 & \textbf{0.83} \\
EM        & None & 19.6 & 14.06 & 10.4  & 5.19 & 3.37 & 2.73 & 3.01 & 1.45 & 1.03 \\
\LESA     & None & 17.7 & 12.62 & 9.41  & \textbf{4.75} & 2.83 & 2.43 & 2.65 & \textbf{1.08} & 0.89 \\
\hline
RLLS & TS   & \textbf{6.80}   & \textbf{5.03}  & \textbf{2.70}  & 6.01 & 2.45 & \textbf{1.18} & 4.02 & 1.54 & 0.79 \\
EM   & TS   & 13.45  & 9.53  & 6.04  & 5.52 & 2.10 & 1.49 & 3.86 & 1.39 & 0.84 \\
\LESA & TS  & 12.58  & 7.57  & 5.39  & \textbf{5.06} & \textbf{1.40} & 1.33 & \textbf{3.25} & \textbf{0.74} & \textbf{0.71} \\
\hline
RLLS & BCTS  & 7.44  & 4.66  & 2.78 & 4.50 & 1.47 & 0.83 & 4.53 & 1.38 & 0.79 \\
EM   & BCTS  & 4.97  & 3.16  & 2.81 & 3.72 & 1.23 & 0.87 & 3.94 & 1.15 & 0.81 \\
\LESA & BCTS & \textbf{3.83}  & \textbf{2.07}  & \textbf{2.39}  & \textbf{2.28} & \textbf{0.57} & \textbf{0.68} & \textbf{2.12} & \textbf{0.33} & \textbf{0.41} \\
\hline
RLLS  & NBVS & 7.39  & 4.12  & 2.74  & 6.33 & 2.24 & \textbf{1.23} & 4.34 & 1.32 & 0.84 \\
EM    & NBVS & 8.66  & 3.57  & 3.07  & 4.89 & 1.53 & \textbf{1.23} & 3.98 & 1.08 & 0.83 \\
\LESA & NBVS & \textbf{8.29}  & \textbf{3.25}  & \textbf{2.93}  & \textbf{4.63} & \textbf{1.5}  & 1.4 & \textbf{3.35} & \textbf{0.66} & \textbf{0.73} \\
\hline
RLLS & VS   & 7.44  & 4.07  & 2.81  & 6.47 & 2.17 & 1.22 & 4.57 & 1.26 & 0.81 \\
EM   & VS   & 2.83  & 0.89  & 1.03  & 4.53 & 1.12 & 0.97 & 4.25 & 1.03 & 0.78 \\
\LESA & VS  & \textbf{2.32}  & \textbf{0.78}  & \textbf{0.96}  & \textbf{2.54} & \textbf{0.54} & \textbf{0.59} & \textbf{2.20} & \textbf{0.35} & \textbf{0.34} \\
\hline
\end{tabular}
\vspace{6pt}
\caption{\textbf{CIFAR 10: Comparison of CC, RLLS, EM and \LESA under Dirichlet shift} Values reported are MSE between the estimated shift weight and the true weight. All numbers are reported on the scale of $10^{-3}$ and are means across 50 runs for each $\alpha$. Weight for EM and \LESA are calculated by dividing the estimated target label distribution by source label distribution calculated with $\frac{1}{N_{v}}\sum_{k=0}^{k=N_{v}}p(y_{i}|\mathbf{x}_{k})$ for each class over the validation set to be consistent with previous work \cite{alexandari2020maximum}. For CC, the source distribution for weight is simply the class distribution of the validation set. Numbers in bold indicate the best-performing method in the group.}
\label{Table1}
\end{table*}

We evaluate the performance of \LESA on CIFAR10 \cite{krizhevsky2014cifar} and MNIST \cite{lecun2010mnist}. For comparison, we choose three baselines: EM, RLLS, and a naive classify-and-count approach. In previous work, EM has been established as the best-performing post hoc label shift estimation method~\cite{alexandari2020maximum, garg_unified_2020}. RLLS is a regularized version of BBSL \cite{lipton_detecting_2018}, empirically shown to outperform or perform on par with BBSL. We use it as the method representing the confusion matrix approaches \footnote{For RLLS, the two parameters $\alpha$ and $\delta$ are set to be $0.01$ and $0.05$ consistent to the original RLLS implementation}. RLLS-Hard is a version of RLLS that uses the top prediction instead of the posterior for label shift estimation. The classify-and-count (CC) approach is a naive method for label shift estimation, where the estimated $\hat{p}_t(y)$ is simply the label distribution of $argmax(\hat{p}(y|\mathbf{x}_{k}))$.

To simulate label shift, we take the same approach as the previous work \cite{alexandari2020maximum, lipton_detecting_2018, azizzadenesheli_regularized_2019}, which performs Dirichlet shift on the test set. Dirichlet shift generates the testing set class distribution by sampling from Dirichlet distributions of different parameters $\boldsymbol{\alpha} = [\alpha]_{m}$, where $m$ is the number of classes and $\alpha > 0$. A smaller $\alpha$ results in a more extreme class distribution shift. The Dirichlet shift-generated test set label distribution is then sampled from the testing set without replacement with the largest possible testing set size.  To ensure the robustness of the result, for each $\alpha$, we perform 50 instances of Dirichlet shift, each with a different random state, resulting in different generated target distributions.

\begin{figure*}[h]
    \centering

    \subfloat[\small{CIFAR10: Calibration with BCTS}]{
        \includegraphics[width = .38\textwidth]{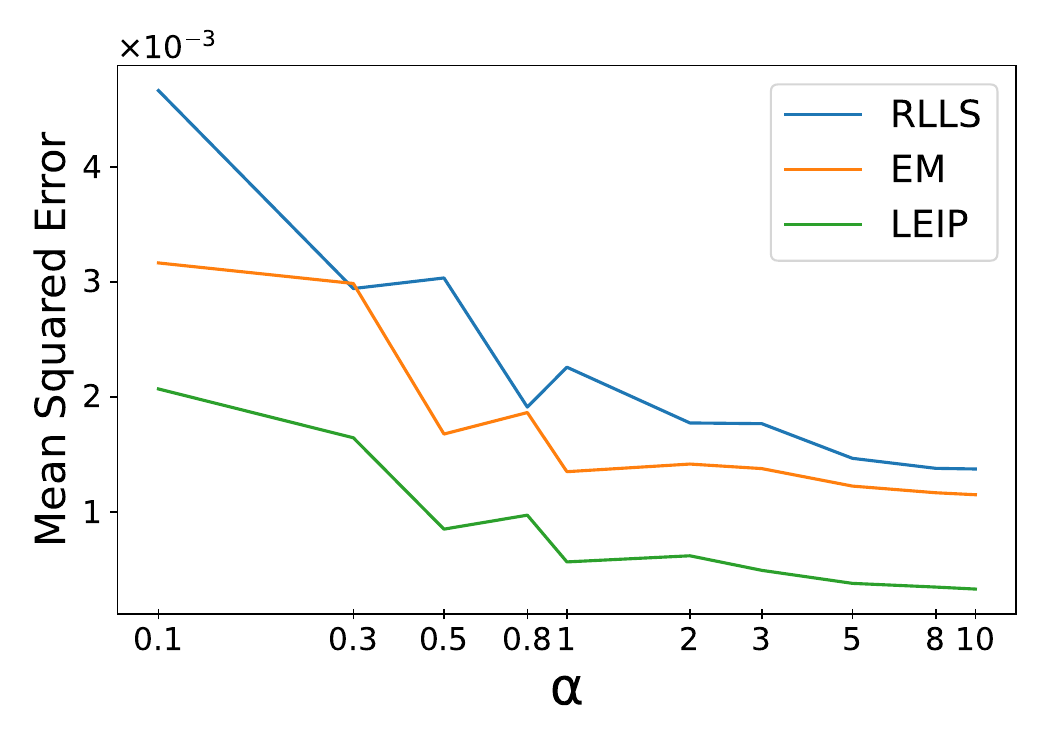} 
        \label{fig:CIFAR_a}
    }
    \hspace{20pt}
    \subfloat[\small{CIFAR10: Calibration with VS}]{
        \includegraphics[width = .38\textwidth]{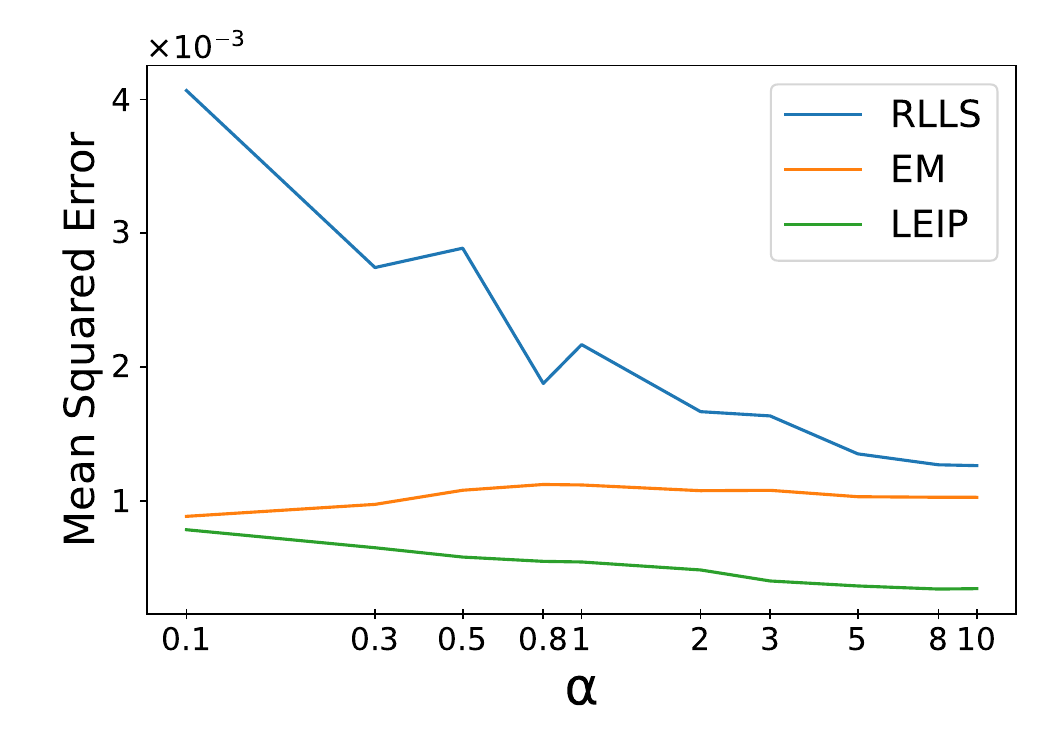} 
        \label{fig:CIFAR_b}
    }

     \subfloat[\small{MNIST: Calibration with BCTS}]{
        \includegraphics[width = .38\textwidth]{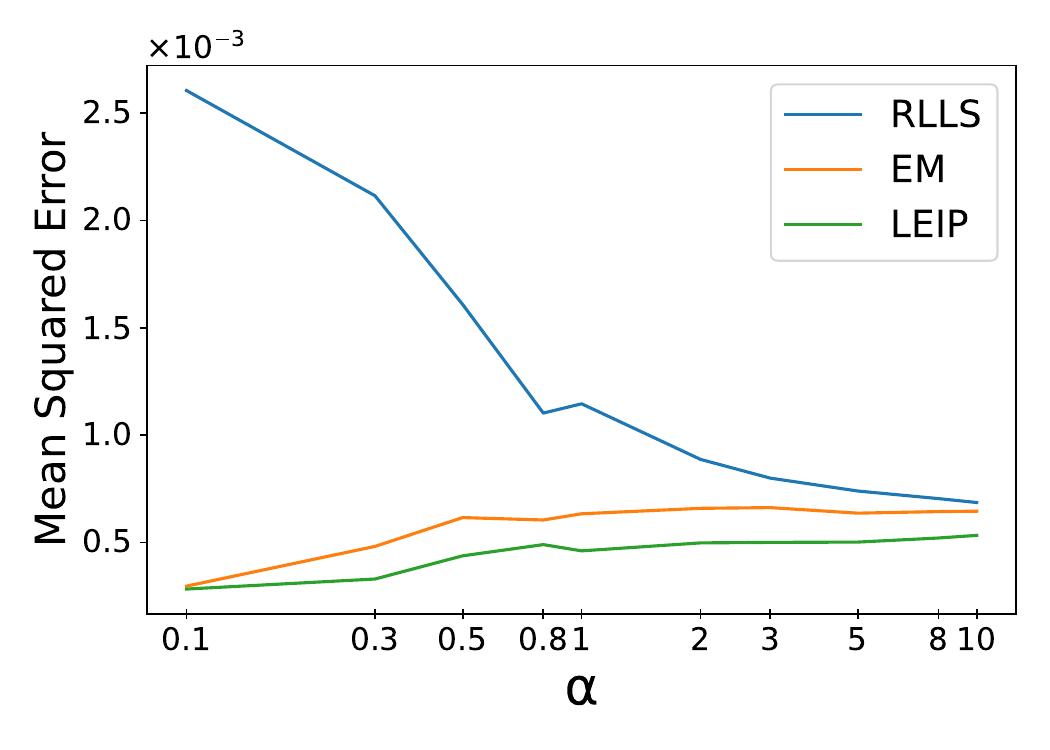} 
        \label{fig:MNIST_a}
    }
    \hspace{20pt}
    \subfloat[\small{MNIST: Calibration with VS}]{
        \includegraphics[width = .38\textwidth]{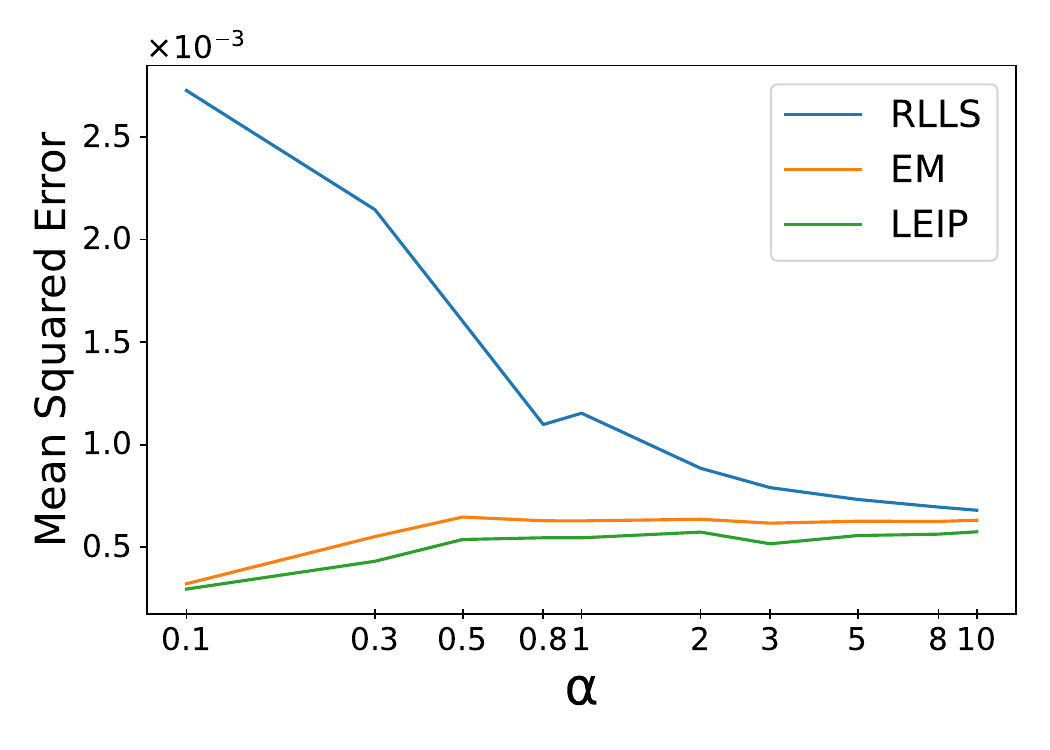} 
        \label{fig:MNIST_b}
    }

    \caption{Comparison of CC, RLLS, EM and \LESA under Dirichlet shift for a wide range of $\alpha$. Numbers reported are on the scale of $10^{-3}$; for each $\alpha$, the MSE reported are average across 50 runs. Validation set size 2500 for CIFAR10 and 1500 for MNIST.}
    \label{fig:Comparison}
\end{figure*}




Previous work has demonstrated different approaches to sampling the Dirichlet-shifted test set. Some studies sample the target distribution with replacement \cite{alexandari2020maximum, lipton_detecting_2018} so that a set number of testing samples can be achieved, while others do not \cite{azizzadenesheli_regularized_2019} and the sampled testing set size would be the maximum size for the given target distribution without replacement. Sampling with replacement can lead to issues, particularly in cases of extreme label shift or when the test set has a low number of samples per class. This method can repeatedly sample the same instance, leading to less robust results and unrealistic distributions. Therefore, we choose to sample without replacement in line with \cite{azizzadenesheli_regularized_2019}.

For metrics of evaluating label shift estimation, we take the same approach as the previous work \cite{alexandari2020maximum, lipton_detecting_2018, azizzadenesheli_regularized_2019} that report the mean squared error between the true class shift weight and the estimated class shift weight. The true class weight is calculated by $w = \frac{p_t(y)}{p_s(y)}$ where the $p_s(y)$ is the source class distribution and the $p_t(y)$ is the target class distribution. For RLLS \cite{azizzadenesheli_regularized_2019}, it directly outputs the class shift weight estimation, but for EM and \LESA that estimate the shifted label distribution $\hat{p}(y)$, the weight needs to be calculated. In previous work \cite{alexandari2020maximum} the weight is calculated by $\hat{w} = \frac{\hat{p}(y)}{p_{e}(y)}$ where for each $p_{e}(y_{i})$ is calculated as $\frac{1}{N_{v}}\sum_{k=0}^{k=N_{v}}p(y_{i}|\mathbf{x}_{k})$ over the validation set with $N_{v}$ samples. Essentially taking the average classifier output posteriors for each class of the validation set. 

However, in our experiment, we found that for $p_{e}(y)$, simply hard counting the class distribution over the validation set obtains better results for EM and \LESA and makes them perform better than RLLS in the case where the classifier is uncalibrated or paired with a less parameterized calibration method. To be consistent with the previous work \cite{alexandari2020maximum}, in this section, we calculate the $\hat{w}$ for EM and \LESA as suggested by Alexandari et al. \cite{alexandari2020maximum} by setting the $p_{e}(y)$ as the average classifier posteriors for each class of the validation set. The results of setting $p_{e}(y)$ by hard counting on the validation set are available in the Appendix. In both scenarios, \LESA outperforms EM.

Both EM and \LESA require calibration to achieve optimal performance. We follow the calibration approach of Alexandari et al. \cite{alexandari2020maximum}, which compared the performance of label shift estimators under four calibration methods: temperature scaling (TS) \cite{hinton2015distilling}, bias-corrected temperature scaling (BCTS) \cite{alexandari2020maximum}, vector scaling (VS) \cite{guo_calibration_2017}, and no-bias vector scaling (NBVS) \cite{alexandari2020maximum}. Each calibration method is paired with three validation sizes ranging from 5\% to 20\% of the training set size. The validation size significantly influences the quality of calibration and the estimation of the confusion matrix.

\begin{table*}[htbp]
\centering
\small
\setlength{\tabcolsep}{6.5pt}
\begin{tabular}{cc|ccc|ccc|ccc}
\makecell{Shift \\ Estimator}& \makecell{Calibration \\ Method} & \multicolumn{3}{c|}{$\alpha=0.1$} &\multicolumn{3}{|c|}{$\alpha=1$}&\multicolumn{3}{|c}{$\alpha=10$}\\
\hline
\multicolumn{2}{c}{Validation Size} & 500 & 1500 & 2000 & 500 & 1500 & 2000 & 500 & 1500 & 2000 \\
\hline
CC        & None & 13.3  & 12.0  & 11.5  & \textbf{2.74} & 2.49 & 2.32 & \textbf{0.60} & \textbf{0.48} & \textbf{0.45} \\
RLLS-Hard & None & 13.1  & \textbf{2.58}  & 2.32  & 7.43 & 1.06 & 1.04 & 4.89 & 0.57 & 0.82 \\
RLLS      & None & \textbf{10.8}  & 2.62  & \textbf{2.29}  & 7.39 & 1.12 & \textbf{0.97} & 5.35 & 0.65 & 0.64 \\
EM        & None & 13.9  & 2.99  & 3.20  & 7.92 & 1.03 & 1.16 & 5.49 & 0.66 & 0.65 \\
\LESA     & None & 14.2  & 3.14  & 3.29  & 8.00 & \textbf{0.97} & 1.14 & 5.46 & 0.67 & 0.67 \\
\hline
RLLS & TS   & \textbf{9.52}  & \textbf{2.61}  & \textbf{2.52}  & \textbf{6.91} & \textbf{1.22} & \textbf{1.13} & \textbf{5.15} & \textbf{0.72} & \textbf{0.74} \\
EM   & TS   & 13.5  & 3.50  & 3.74  & 7.63 & 1.24 & 1.39 & 5.32 & 0.74 & 0.77 \\
\LESA & TS  & 13.9  & 3.74  & 3.79  & 8.01 & 1.49 & 1.57 & 5.38 & 1.00 & 0.91 \\
\hline
RLLS  & BCTS & 10.1  & 2.60  & 2.58  & 6.08 & 1.15 & 1.01 & 4.32 & 0.69 & 0.63 \\
EM    & BCTS & 1.40  & 0.30  & 0.32  & 3.60 & 0.63 & 0.64 & 3.99 & 0.65 & 0.59 \\
\LESA & BCTS & \textbf{0.97}  & \textbf{0.28}  & \textbf{0.26}  & \textbf{2.60} & \textbf{0.46} & \textbf{0.44} & \textbf{3.02} & \textbf{0.53} & \textbf{0.44} \\
\hline
RLLS  & NBVS & 9.73  & 2.78  & 2.74  & 6.59 & 1.18 & 1.08 & 4.73 & 0.68 & 0.70 \\
EM    & NBVS & 3.39  & 0.81  & 0.40  & 4.62 & 0.61 & 0.62 & 4.74 & 0.64 & 0.65 \\
\LESA & NBVS & \textbf{3.13}  & \textbf{0.76}  & \textbf{0.36}  & \textbf{3.43} & \textbf{0.55} & \textbf{0.44} & \textbf{3.59} & \textbf{0.53} & \textbf{0.48} \\
\hline
RLLS & VS   & 11.58 & 2.73  & 2.71  & 5.82 & 1.15 & 1.10 & 3.79 & 0.68 & 0.73 \\
EM   & VS   & 4.52  & 0.32  & 0.28  & 4.16 & 0.63 & 0.66 & 3.57 & 0.63 & 0.68 \\
\LESA & VS  &  \textbf{4.20}  &  \textbf{0.30}  &  \textbf{0.23}  &  \textbf{3.37} &  \textbf{0.55} &  \textbf{0.46} &  \textbf{2.92} &  \textbf{0.58} &  \textbf{0.50} 
\end{tabular}
\vspace{6pt}
\caption{\textbf{MNIST: Comparison of CC, RLLS, EM and \LESA under Dirichlet shift} Values reported are MSE between the estimated shift weight and the true weight. All numbers are reported on the scale of $10^{-3}$ and are means across 50 runs for each $\alpha$. The same settings for calculating weight and source distribution are used as introduced in the caption of Table \ref{Table1}.}
\label{Table2}
\end{table*}


For CIFAR-10, the train/validation/test split is 20,000/10,000/30,000, and for MNIST, it is 10,000/10,000/40,000. For CIFAR-10, we train a ResNet-18 \cite{He_2016_CVPR} for 200 epochs. For MNIST, we train a two-layer Multilayer Perceptron (MLP) with ReLU activation and 256 hidden units on the training set for five epochs. For each $\alpha$, we perform 50 instances of Dirichlet shift, each with a different random state, resulting in different generated target distributions. The reported MSE is the average across 50 instances of Dirichlet shift. 

Tables \ref{Table1} and \ref{Table2} present the results of comparisons under different calibration methods, validation sizes, and $\alpha$ values. When the classifier is uncalibrated or calibrated with non-bias-corrected methods such as temperature scaling, the confusion matrix-based method RLLS outperforms both EM and \LESA. This result is consistent with the findings of Alexandari et al. \cite{alexandari2020maximum}. However, when the classifier is paired with bias-corrected calibration methods such as BCTS or VS, \LESA outperforms all other methods. The best performance is achieved by combining \LESA with BCTS or VS.


Figures \ref{fig:Comparison} show the results on CIFAR-10 and MNIST for various $\alpha$ values under the BCTS and VS calibration. We can observe that, under the calibration of BCTS or VS, \LESA consistently outperforms EM and RLLS across a wide range of $\alpha$, representing varying intensities of label shift.

Another metric that the previous work often presents for evaluating label shift estimation is the change in accuracy after adapting the classifier with the estimated weight $w$ \cite{lipton_detecting_2018, alexandari2020maximum}. To adapt the classifier with the estimated weight $w$, two approaches have been taken by previous work: either retraining the classifier with importance weighting \cite{lipton_detecting_2018} or updating the posterior of the test set with the estimated prior using Equation \ref{LESA:1} \cite{alexandari2020maximum}. However, importance weighting has been proven ineffective for deep neural networks, as its effect diminishes in later epochs \cite{byrd_what_2019}. For the approach of updating each sample's posterior with the estimated weight $w$ when $\alpha$ is 0.1, we observe a slight increase in accuracy when adapting the posteriors using Equation \ref{LESA:1} with the $\frac{\hat{p}^{(a)}(y_{j})}{p(y_{j})}$ substituted by weight estimated by \LESA EM or RLLS. However, for larger values of $\alpha$, the difference in accuracy is not significant. 


In the case of $\alpha = 0.1$, the label shift is extreme, resulting in many minority classes and even some classes having no representation in the testing set. In this scenario, when one or a few dominating majority classes are present in the testing set, it is much easier to achieve a high accuracy or a high average precision by predicting bias towards the majority classes, and the performance of the minority class would have little effect on the accuracy making accuracy an unfair metric for evaluation. Instead, macro-averaged recall seems more appropriate, giving more weight to minority classes. Although we saw an increase in accuracy and average precision, The average recall drops by a significant margin for $\alpha = 0.1$ compared to the un-adapted classifier. This means that by posing a new significantly shifted prior on the posterior, although we can increase the average precision and accuracy by decreasing the false positive for minority classes, however in exchange, we sacrifice the number of true positives for the minority class that causes more in the false negative and thus lower macro average recall, which is a more important metric for measuring performance in the case of many minority classes present.

The key takeaway is that the current approaches for adapting classifiers to new priors are not consistently effective in improving classifier performance. Consequently, there is a need to develop more robust methods in future research. Theoretically, more precise label shift estimation should lead to better adaptation of classifiers. In this context, \LESA has consistently demonstrated superior performance in label shift estimation. The estimated label distribution has broader applications beyond classifier adaptation, including Quantification \cite{gonzalez_review_2017, forman2008quantifying} and PU-learning \cite{elkan2008learning}.

\section{Conclusion}
In conclusion, we proposed \LESA, a novel label shift estimation method that performs incremental prior updates on test sets with unknown label distributions. \LESA is motivated by three intuitive assumptions natural for label shift estimation and most modern probabilistic classifiers. Based on these assumptions, \LESA employs an approach where a rough class estimation is obtained by isolating a set of samples that are unlikely to be incorrect, and it performs prior updates incrementally and in a non-aggregative manner on the rest of the samples where prior updates would make a difference in the top label. \LESA does not require any hyperparameter tuning and relies on a weaker notion of calibration compared to previous work. Empirical testing shows that \LESA consistently outperforms current state-of-the-art methods on reproducible benchmarks set up within the guidance of prior works.

For future work, it is crucial to address the task of label shift estimation in a streaming setup \cite{baby2023online}. Additionally, as discussed in the experiment section, current methods for adjusting classifiers to new label distributions are ineffective and yield non-robust results, necessitating further research. Furthermore, we will extend our approach to the relaxed label shift problem \cite{garg_rlsbench_2023, tachet2020domain}.

\newpage

\bibliographystyle{siam}
\bibliography{reference}

@inproceedings{elkan2008learning,
  title={Learning classifiers from only positive and unlabeled data},
  author={Elkan, Charles and Noto, Keith},
  booktitle={Proceedings of the 14th ACM SIGKDD international conference on Knowledge discovery and data mining},
  pages={213--220},
  year={2008}
}

@article{gonzalez_review_2017,
	title = {A {Review} on {Quantification} {Learning}},
	volume = {50},
	issn = {0360-0300},
	url = {https://doi.org/10.1145/3117807},
	doi = {10.1145/3117807},
	number = {5},
	urldate = {2023-02-20},
	journal = {ACM Comput. Surv.},
	author = {González, Pablo and Castaño, Alberto and Chawla, Nitesh V. and Coz, Juan José Del},
	month = sep,
	year = {2017},
	pages = {74:1--74:40},
}

@article{saerens_adjusting_2002,
	title = {Adjusting the {Outputs} of a {Classifier} to {New} a {Priori} {Probabilities}: {A} {Simple} {Procedure}},
	volume = {14},
	issn = {0899-7667},
	shorttitle = {Adjusting the {Outputs} of a {Classifier} to {New} a {Priori} {Probabilities}},
	url = {https://doi.org/10.1162/089976602753284446},
	doi = {10.1162/089976602753284446},
	number = {1},
	urldate = {2023-03-05},
	journal = {Neural Computation},
	author = {Saerens, Marco and Latinne, Patrice and Decaestecker, Christine},
	month = jan,
	year = {2002},
	pages = {21--41},
}

@inproceedings{
azizzadenesheli_regularized_2019,
title={Regularized Learning for  Domain Adaptation under Label Shifts},
author={Kamyar Azizzadenesheli and Anqi Liu and Fanny Yang and Animashree Anandkumar},
booktitle={International Conference on Learning Representations},
year={2019},
url={https://openreview.net/forum?id=rJl0r3R9KX},
}

@inproceedings{lipton_detecting_2018,
  title={Detecting and correcting for label shift with black box predictors},
  author={Lipton, Zachary and Wang, Yu-Xiang and Smola, Alexander},
  booktitle={International conference on machine learning},
  pages={3122--3130},
  year={2018},
  organization={PMLR}
}

@inproceedings{alexandari2020maximum,
  title={Maximum likelihood with bias-corrected calibration is hard-to-beat at label shift adaptation},
  author={Alexandari, Amr and Kundaje, Anshul and Shrikumar, Avanti},
  booktitle={International Conference on Machine Learning},
  year={2020},
  organization={PMLR}
}

@inproceedings{garg_unified_2020,
	title = {A {Unified} {View} of {Label} {Shift} {Estimation}},
	volume = {33},
	url = {https://proceedings.neurips.cc/paper/2020/hash/219e052492f4008818b8adb6366c7ed6-Abstract.html},
	urldate = {2024-03-30},
	booktitle = {Advances in {Neural} {Information} {Processing} {Systems}},
	publisher = {Curran Associates, Inc.},
	author = {Garg, Saurabh and Wu, Yifan and Balakrishnan, Sivaraman and Lipton, Zachary},
	year = {2020},
	pages = {3290--3300},
}

@misc{garg_rlsbench_2023,
	title = {{RLSbench}: {Domain} {Adaptation} {Under} {Relaxed} {Label} {Shift}},
	shorttitle = {{RLSbench}},
	url = {http://arxiv.org/abs/2302.03020},
	urldate = {2024-04-09},
	publisher = {arXiv},
	author = {Garg, Saurabh and Erickson, Nick and Sharpnack, James and Smola, Alex and Balakrishnan, Sivaraman and Lipton, Zachary C.},
	month = jun,
	year = {2023},
	note = {arXiv:2302.03020 [cs, stat]},
}

@inproceedings{zhang2013domain,
  title={Domain adaptation under target and conditional shift},
  author={Zhang, Kun and Sch{\"o}lkopf, Bernhard and Muandet, Krikamol and Wang, Zhikun},
  booktitle={International conference on machine learning},
  pages={819--827},
  year={2013},
  organization={Pmlr}
}

@inproceedings{guo_calibration_2017,
	title = {On {Calibration} of {Modern} {Neural} {Networks}},
	url = {https://proceedings.mlr.press/v70/guo17a.html},
	language = {en},
	urldate = {2023-10-11},
	booktitle = {Proceedings of the 34th {International} {Conference} on {Machine} {Learning}},
	publisher = {PMLR},
	author = {Guo, Chuan and Pleiss, Geoff and Sun, Yu and Weinberger, Kilian Q.},
	month = jul,
	year = {2017},
	note = {ISSN: 2640-3498},
	pages = {1321--1330},
}

@article{silva_filho_classifier_2023,
	title = {Classifier calibration: a survey on how to assess and improve predicted class probabilities},
	volume = {112},
	issn = {1573-0565},
	shorttitle = {Classifier calibration},
	url = {https://doi.org/10.1007/s10994-023-06336-7},
	doi = {10.1007/s10994-023-06336-7},
	language = {en},
	number = {9},
	urldate = {2023-11-23},
	journal = {Mach Learn},
	author = {Silva Filho, Telmo and Song, Hao and Perello-Nieto, Miquel and Santos-Rodriguez, Raul and Kull, Meelis and Flach, Peter},
	month = sep,
	year = {2023},
	pages = {3211--3260},
}

@inproceedings{kull_beyond_2019,
	title = {Beyond temperature scaling: {Obtaining} well-calibrated multi-class probabilities with {Dirichlet} calibration},
	volume = {32},
	shorttitle = {Beyond temperature scaling},
	url = {https://proceedings.neurips.cc/paper/2019/hash/8ca01ea920679a0fe3728441494041b9-Abstract.html},
	urldate = {2024-01-03},
	booktitle = {Advances in {Neural} {Information} {Processing} {Systems}},
	publisher = {Curran Associates, Inc.},
	author = {Kull, Meelis and Perello Nieto, Miquel and Kängsepp, Markus and Silva Filho, Telmo and Song, Hao and Flach, Peter},
	year = {2019},
}

@book{mclachlan2005discriminant,
  title={Discriminant analysis and statistical pattern recognition},
  author={McLachlan, Geoffrey J},
  year={1992},
  publisher={John Wiley \& Sons}
}

@inproceedings{latinne2001adjusting,
  title={Adjusting the outputs of a classifier to new a priori probabilities may significantly improve classification accuracy: Evidence from a multi-class problem in remote sensing},
  author={Latinne, Patrice and Saerens, Marco and Decaestecker, Christine},
  booktitle={ICML},
  volume={1},
  pages={298--305},
  year={2001}
}

@article{krizhevsky2014cifar,
  title={The CIFAR-10 dataset},
  author={Krizhevsky, Alex and Nair, Vinod and Hinton, Geoffrey and others},
  journal={online: http://www. cs. toronto. edu/kriz/cifar. html},
  volume={55},
  number={5},
  pages={2},
  year={2014}
}

@article{lecun2010mnist,
  title={MNIST handwritten digit database},
  author={LeCun, Yann and Cortes, Corinna and Burges, CJ},
  journal={ATT Labs [Online]. Available: http://yann.lecun.com/exdb/mnist},
  volume={2},
  year={2010}
}

@InProceedings{He_2016_CVPR,
author = {He, Kaiming and Zhang, Xiangyu and Ren, Shaoqing and Sun, Jian},
title = {Deep Residual Learning for Image Recognition},
booktitle = {Proceedings of the IEEE Conference on Computer Vision and Pattern Recognition (CVPR)},
month = {June},
year = {2016}
}

@article{hinton2015distilling,
  title={Distilling the knowledge in a neural network},
  author={Hinton, Geoffrey and Vinyals, Oriol and Dean, Jeff},
  journal={arXiv preprint arXiv:1503.02531},
  year={2015}
}

@article{fang2020rethinking,
  title={Rethinking importance weighting for deep learning under distribution shift},
  author={Fang, Tongtong and Lu, Nan and Niu, Gang and Sugiyama, Masashi},
  journal={Advances in neural information processing systems},
  volume={33},
  pages={11996--12007},
  year={2020}
}

@inproceedings{byrd_what_2019,
  title={What is the effect of importance weighting in deep learning?},
  author={Byrd, Jonathon and Lipton, Zachary},
  booktitle={International conference on machine learning},
  year={2019},
  organization={PMLR}
}

@INBOOK{6279353,
  author={Quiñonero-Candela, Joaquin and Sugiyama, Masashi and Schwaighofer, Anton and Lawrence, Neil D.},
  booktitle={Dataset Shift in Machine Learning}, 
  title={Covariate Shift by Kernel Mean Matching}, 
  year={2009},
  volume={},
  number={},
  pages={131-160},
  doi={}}

@inproceedings{guo2020ltf,
  title={LTF: A Label Transformation Framework for Correcting Label Shift},
  author={Guo, Jiaxian and Gong, Mingming and Liu, Tongliang and Zhang, Kun and Tao, Dacheng},
  booktitle={International Conference on Machine Learning},
  pages={3843--3853},
  year={2020},
  organization={PMLR}
}

@article{platt1999probabilistic,
  title={Probabilistic outputs for support vector machines and comparisons to regularized likelihood methods},
  author={Platt, John and others},
  journal={Advances in large margin classifiers},
  year={1999},
  publisher={Cambridge, MA}
}

@article{kim2020distribution,
  title={Distribution aligning refinery of pseudo-label for imbalanced semi-supervised learning},
  author={Kim, Jaehyung and Hur, Youngbum and Park, Sejun and Yang, Eunho and Hwang, Sung Ju and Shin, Jinwoo},
  journal={Advances in neural information processing systems},
  year={2020}
}

@inproceedings{wei2021crest,
  title={Crest: A class-rebalancing self-training framework for imbalanced semi-supervised learning},
  author={Wei, Chen and Sohn, Kihyuk and Mellina, Clayton and Yuille, Alan and Yang, Fan},
  booktitle={Proceedings of the IEEE/CVF conference on computer vision and pattern recognition},
  pages={10857--10866},
  year={2021}
}

@inproceedings{
zhu2023generalized,
title={Generalized Logit Adjustment: Calibrating Fine-tuned Models by Removing Label Bias in Foundation Models},
author={Beier Zhu and Kaihua Tang and Qianru Sun and Hanwang Zhang},
booktitle={Advances in neural information processing systems},
year={2023},
url={https://openreview.net/forum?id=9qG6cMGUWk}
}

@article{forman2008quantifying,
  title={Quantifying counts and costs via classification},
  author={Forman, George},
  journal={Data Mining and Knowledge Discovery},
  volume={17},
  pages={164--206},
  year={2008},
  publisher={Springer}
}

@inproceedings{tachet2020domain,
      title={Domain Adaptation with Conditional Distribution Matching and Generalized Label Shift},
      author={Tachet des Combes, Remi and Zhao, Han and Wang, Yu-Xiang and Gordon, Geoff},
      year={2020},
      booktitle={Advances in Neural Information Processing Systems}
}

@inproceedings{
baby2023online,
title={Online Label Shift: Optimal Dynamic Regret meets Practical Algorithms},
author={Dheeraj Baby and Saurabh Garg and Tzu-Ching Yen and Sivaraman Balakrishnan and Zachary Chase Lipton and Yu-Xiang Wang},
booktitle={Thirty-seventh Conference on Neural Information Processing Systems},
year={2023},
url={https://openreview.net/forum?id=Ki6DqBXss4}
}
\onecolumn
\appendixsiam
\renewcommand{\theHsection}{A\arabic{section}}
\section{Appendix} 
Here we present the appendix accompanying the main article.
\section{Derivation of Equation \ref{LESA:1} from Bayes' Theorem} \label{App1}
The main article Equation 3 is a well-known formula derived from the Bayes' theorem \cite{saerens_adjusting_2002}. For a probabilistic classifier $\mathcal{F}$ that provides a score to a sample $x$ from a source distribution $s$ according to Bayes' theorem:

\begin{equation}
p_s(y|x) = \frac{p_s(x|y)p_s(y)}{p_s(x)}
\label{ap:-1}
\end{equation}

After label shift, suppose we encounter a target distribution $t$ where $p_t(y) \neq p_s(y)$, leading to a change in the overall feature distribution, $p_t(x) \neq p_s(x)$. The classifier $\mathcal{F}$ gives the new posterior:

\begin{equation}
p_t(y|x) = \frac{p_t(x|y)p_t(y)}{p_t(x)}
\label{ap:0}
\end{equation}

Equation \ref{ap:-1} and \ref{ap:0} can be re-arrange into

\begin{equation}
p_t(x|y) = p_t(y|x) \frac{p_t(x)}{p_t(y)}
\label{ap:1}
\end{equation}

\begin{equation}
p_s(x|y) = p_s(y|x) \frac{p_s(x)}{p_s(y)}
\label{ap:2}
\end{equation}

Since $p_t(y|x) = p_s(y|x)$ by by equating Equation \ref{ap:1} and \ref{ap:2} also defining $f(x) = \frac{p_s(x)}{p_t(x)}$, we find:

\begin{equation}
p_t(y|x) = f(x) \frac{p_t(y)}{p_s(y)} p_s(y|x)
\end{equation}

Since $\sum_{j=1}^{m} p_t(y_j|x) = 1$, we obtain:

\begin{equation}
f(x) = \left[ \sum_{j=1}^{m} \frac{p_t(y_j)}{p_s(y_j)} p_s(y_j|x) \right]^{-1},
\end{equation}

and consequently:

\begin{equation}
p_t(y_i|x) = \frac{\frac{p_t(y_i)}{p_s(y_i)} p_s(y_i|x)}{\sum\limits_{j=1}^{m} \frac{p_t(y_j)}{p_s(y_j)} p_s(y_j|x)}  
\end{equation}

Which is Equation \ref{LESA:1} in the main article.

\newpage
\section{Dirichlet Shift examples} \label{App2}
In this section, we present the label distribution generated by Dirichlet Shift for various values of $\alpha$. Smaller values of $\alpha$ result in more extreme cases of label shift, whereas larger values of $\alpha$ yield label distributions closer to class-balanced distributions. Table \ref{tab:dirichlet_shift} illustrates examples of class distributions generated by Dirichlet Shift with six different values of $\alpha$ for a ten-class problem.

\begin{table}[htbp]
    \centering
    \small
    \begin{tabular}{c|cccccccccc}
        \toprule
        \(\alpha\) & Class 1 & Class 2 & Class 3 & Class 4 & Class 5 & Class 6 & Class 7 & Class 8 & Class 9 & Class 10 \\
        \midrule
        0.1 & 0.008 & 0.003 & 0.239 & 0.144 & 0.000 & 0.000 & 0.002 & 0.443 & 0.161 & 0.000\\
        0.5 & 0.010 & 0.084 & 0.001 & 0.693 & 0.011 & 0.001 & 0.043 & 0.025 & 0.057 & 0.074 \\
        1   & 0.025 & 0.066 & 0.118 & 0.002 & 0.084 & 0.020 & 0.015 & 0.464 & 0.186 & 0.020 \\
        3   & 0.087 & 0.038 & 0.146 & 0.060 & 0.098 & 0.165 & 0.063 & 0.203 & 0.091 & 0.049 \\
        5   & 0.122 & 0.071 & 0.116 & 0.188 & 0.087 & 0.085 & 0.106 & 0.094 & 0.047 & 0.084 \\
        10  & 0.115 & 0.088 & 0.102 & 0.058 & 0.104 & 0.102 & 0.130 & 0.102 & 0.070 & 0.130 \\
        \bottomrule
    \end{tabular}
    \vspace{5pt}
    \caption{Label distribution generated by Dirichlet Shift for different \(\alpha\) values each row sums to 1}
    \label{tab:dirichlet_shift}
\end{table}

\section{Comparison of different approaches of calculating weight for EM and \LESA}\label{App3}

\begin{table}[!htbp]
\centering
\small
\setlength{\tabcolsep}{6.5pt}
\begin{tabular}{cc|ccc|ccc|ccc}
\makecell{Shift \\ Estimator}& \makecell{Calibration \\ Method} & \multicolumn{3}{c|}{$\alpha=0.1$} &\multicolumn{3}{|c|}{$\alpha=1$} & \multicolumn{3}{|c}{$\alpha=10$}\\
\hline
\multicolumn{2}{c}{Validation Size} & 1000 & 2500 & 4000& 1000 & 2500 & 4000 & 1000 & 2500 & 4000 \\
\hline
CC        & None & 53.2 & 50.1 & 49.6 & 9.78 & 9.23 & 9.09 & 1.41 & 1.30 & 1.29\\
RLLS-Hard & None & 6.68 & 3.85 & 2.61 & 5.02 & 2.03 & 1.58 & 2.66 & 1.15 & 1.10 \\
RLLS      & None & 6.07 & 5.27 & 2.60 & 5.42 & 2.55 & 1.22 & 3.38 & 1.62 & 0.83 \\
EM        & None & 10.0 & 8.43 & 8.65 & 2.61 & 2.37 & 2.42 & 0.66 & 0.49 & 0.53 \\
\LESA      & None & 8.26 & 7.21 & 7.64 & 2.37 & 2.2  & 2.17 & 0.57 & 0.55 & 0.53 \\
\hline
RLLS & TS   & 6.80 & 5.03 & 2.7  & 6.01 & 2.45 & 1.18 & 4.02 & 1.54 & 0.79 \\
EM   & TS   & 2.68 & 2.47 & 2.48 & 1.02 & 0.81 & 0.91 & 0.58 & 0.30 & 0.35 \\
\LESA & TS  & 2.05 & 1.97 & 2.26 & 0.66 & 0.66 & 0.67 & 0.33 & 0.34 & 0.33 \\
\hline
RLLS & BCTS & 7.44 & 4.66 & 2.78 & 6.45 & 2.26 & 1.19 & 4.53 & 1.38 & 0.79 \\
EM   & BCTS & 4.97 & 3.17 & 2.81 & 3.91 & 1.35 & 1.09 & 3.94 & 1.15 & 0.81 \\
\LESA & BCTS & 3.82 & 2.07 & 2.39 & 2.28 & 0.57 & 0.68 & 2.11 & 0.33 & 0.41 \\
\hline
RLLS  & NBVS & 7.39 & 4.12 & 2.74 & 6.33 & 2.24 & 1.23 & 4.34 & 1.32 & 0.84 \\
EM    & NBVS & 3.78 & 2.74 & 2.18 & 2.70 & 2.18 & 1.40 & 2.61 & 2.08 & 1.02 \\
\LESA & NBVS & 3.13 & 1.73 & 1.81 & 1.63 & 0.86 & 0.67 & 1.52 & 0.68 & 0.40  \\
\hline
RLLS & VS   & 7.44 & 4.07 & 2.81 & 6.47 & 2.17 & 1.22 & 4.57 & 1.26 & 0.81 \\
EM   & VS   & 2.83 & 0.89 & 1.03 & 4.54 & 1.12 & 0.97 & 4.25 & 1.03 & 0.78 \\
\LESA & VS  & 2.32 & 0.79 & 0.96 & 2.54 & 0.54 & 0.59 & 2.20 & 0.35 & 0.34
\end{tabular}
\vspace{5pt}
\caption{\textbf{CIFAR10: Comparison of CC, RLLS, EM, and \LESA under Dirichlet shift}. Values reported are MSE between the estimated shift weight and the true weight. All numbers are reported on the scale of $10^{-3}$ and are means across 50 runs for each $\alpha$. Weights for EM and \LESA are calculated by dividing the estimated target label distribution by the source label distribution, determined by hard label counts in the validation set. This method, which differs from previous work \cite{alexandari2020maximum}, yields better results for EM and \LESA when the classifier is uncalibrated or calibrated with TS.}
\end{table}

\begin{table}[h!]
\centering
\small
\setlength{\tabcolsep}{6.5pt}
\begin{tabular}{cc|ccc|ccc|ccc}
\makecell{Shift \\ Estimator}& \makecell{Calibration \\ Method} & \multicolumn{3}{c|}{$\alpha=0.1$} &\multicolumn{3}{|c|}{$\alpha=1$}&\multicolumn{3}{|c}{$\alpha=10$}\\
\hline
\multicolumn{2}{c}{Validation Size} & 500 & 1500 & 2000 & 500 & 1500 & 2000 & 500 & 1500 & 2000 \\
\hline
CC        & None & 13.3  & 12.0 & 11.5  & 2.74 & 2.49 & 2.32 & 0.60 & 0.48 & 0.45 \\
RLLS-Hard & None & 13.1  & 2.58  & 2.32  & 7.43 & 1.06 & 1.04 & 4.89 & 0.57 & 0.82 \\
RLLS      & None & 10.8  & 2.62  & 2.29  & 7.39 & 1.12 & 0.97 & 5.35 & 0.65 & 0.64 \\
EM        & None & 0.58  & 0.28  & 0.30  & 1.03 & 0.48 & 0.49 & 1.11 & 0.46 & 0.47 \\ 
\LESA     & None & 0.30  & 0.26  & 0.25  & 0.46 & 0.33 & 0.32 & 0.45 & 0.31 & 0.30 \\
\hline
RLLS & TS   & 9.52  & 2.61  & 2.52  & 6.91 & 1.22 & 1.13 & 5.15 & 0.72 & 0.74 \\
EM   & TS   & 1.25  & 0.40  & 0.43  & 2.17 & 0.81 & 0.73 & 2.39 & 0.75 & 0.73 \\
\LESA & TS  & 0.48  & 0.28  & 0.26  & 0.75 & 0.46 & 0.44 & 0.60 & 0.40 & 0.38 \\
\hline
RLLS & BCTS &  10.1  & 2.60  & 2.58  & 6.08 & 1.15 & 1.01 & 4.32 & 0.69 & 0.63 \\
EM   & BCTS &  1.40  & 0.30  & 0.32  & 3.60 & 0.63 & 0.64 & 3.99 & 0.65 & 0.59 \\
\LESA & BCTS & 0.97  & 0.28  & 0.26  & 2.60 & 0.46 & 0.44 & 3.02 & 0.53 & 0.44 \\
\hline
RLLS & NBVS &  9.73  & 2.78  & 2.74  & 6.59 & 1.18 & 1.08 & 4.73 & 0.68 & 0.70 \\
EM   & NBVS &  1.20  & 0.34  & 0.37  & 3.21 & 0.86 & 0.79 & 3.55 & 0.91 & 0.82 \\
\LESA & NBVS & 0.75  & 0.32  & 0.31  & 1.82 & 0.52 & 0.49 & 2.14 & 0.59 & 0.54 \\
\hline
RLLS & VS   & 11.6 & 2.72  & 2.71  & 5.82 & 1.15 & 1.10 & 3.79 & 0.68 & 0.73 \\
EM   & VS   & 4.53  & 0.32  & 0.28  & 4.16 & 0.63 & 0.66 & 3.57 & 0.63 & 0.68 \\
\LESA & VS  & 4.20  & 0.30  & 0.23  & 3.37 & 0.55 & 0.46 & 2.92 & 0.58 & 0.50
\end{tabular}
\vspace{5pt}
\caption{\textbf{MNIST: Comparison of CC, RLLS, EM, and \LESA under Dirichlet shift}. Values reported are MSE between the estimated shift weight and the true weight. All numbers are reported on the scale of $10^{-3}$ and are means across 50 runs for each $\alpha$. Weights for EM and \LESA are calculated by dividing the estimated target label distribution by the source label distribution, determined by hard label counts in the validation set. This method, which differs from previous work \cite{alexandari2020maximum}, yields better results for EM and \LESA when the classifier is uncalibrated or calibrated with TS.}
\end{table}

\end{document}